\documentclass[10pt,twocolumn,letterpaper]{article}

\usepackage{iccv}
\usepackage{times}
\usepackage{epsfig}
\usepackage{graphicx}
\usepackage{amsmath}
\usepackage{amssymb}

\usepackage{booktabs}
\usepackage{multirow}
\usepackage{color}
\usepackage{textcomp}
\usepackage{float}
\usepackage{authblk}

\usepackage[breaklinks=true,bookmarks=false]{hyperref}

\iccvfinalcopy 

\def\iccvPaperID{1850} 

\ificcvfinal\fi

\begin{document}


\title{Unsupervised High-Resolution Depth Learning From Videos With Dual Networks}

\author[1,\footnote{}]{Junsheng Zhou}
\author[2]{Yuwang Wang}
\author[1]{Kaihuai Qin}
\author[2]{Wenjun Zeng}
\affil[1]{Tsinghua University, Beijing, China \authorcr
{\tt\small zhoujs17@mails.tsinghua.edu.cn, qkh-dcs@mail.tsinghua.edu.cn}}
\affil[2]{Microsoft Research, Beijing, China \authorcr
{\tt\small {yuwwan, wezeng}@microsoft.com}}

\maketitle
\ificcvfinal\thispagestyle{empty}\fi

\footnotetext[1]{Work done as an intern at MSRA.}

\begin{abstract}
Unsupervised depth learning takes the appearance difference between a target view and a view synthesized from its adjacent frame as supervisory signal. Since the supervisory signal only comes from images themselves, the resolution of training data significantly impacts the performance. High-resolution images contain more fine-grained details and provide more accurate supervisory signal. However, due to the limitation of memory and computation power, the original images are typically down-sampled during training, which suffers heavy loss of details and disparity accuracy. In order to fully explore the information contained in high-resolution data, we propose a simple yet effective dual networks architecture, which can directly take high-resolution images as input and generate high-resolution and high-accuracy depth map efficiently. We also propose a Self-assembled Attention (SA-Attention) module to handle low-texture region. The evaluation on the benchmark KITTI and Make3D datasets demonstrates that our method achieves state-of-the-art results in the monocular depth estimation task.
\end{abstract}

\section{Introduction}
\label{sec1}
Estimating depth from RGB images has broad applications such as scene understanding, 3D modelling, robotics, autonomous driving, etc. However, depth estimation from a single image is a well-known ill-posed problem and has inherent scale ambiguity. Recently supervised deep learning methods~\cite{eigen2014depth,liu2014discrete,kuznietsov2017semi} have achieved tremendous success in monocular depth estimation tasks. These methods use convolutional neural networks to predict depth from single RGB image input under the supervision of ground-truth depth obtained by laser scanners. But fully supervised learning methods still suffer from lack of large scale and diverse datasets. So unsupervised methods~\cite{godard2017unsupervised,zhou2017unsupervised} have been proposed to get rid of ground truth labeling and have attracted increasing interest in recent years.

The key idea of unsupervised learning from video is to simultaneously estimate depth of scenes and ego-motion of camera, and use the predicted depth and pose to synthesize target view from source view based on geometric constraint. Then the appearance difference between the synthesized target view and real target view is used as supervisory signal. This unsupervised method is similar to traditional structure-from-motion and stereo matching methods, and therefore suffers from the same problems like occlusion/disocclusion and non-texture region. Since the supervisory signal only comes from images themselves, the quality of images, e.g., the resolution,  significantly impacts the performance. For example, disparity of distant object is always smaller than that of nearby object when the camera is moving forward. Since digital image is a discretized representation of the real world, its resolution limits the accuracy of the disparity. More specifically, it is difficult to distinguish whether the depth of an object is 40m or 80m when its disparity is less than one pixel. Actually we also observed that previous state-of-the-art methods that adopt down-sampling, are prone to produce large error on distant objects. In addition, due to the limitation of memory and computation power, previous methods usually take down-sampled images as training data and predict depth maps with the same size as input. Then the low-resolution results need to be upsampled to the original resolution, resulting in blurred border in this operation. Besides, due to the loss of fine-grained details, slim objects like traffic pole and tree are often neglected, which is a serious safety problem in practical application.

In this paper, we propose a simple yet effective dual (i.e., high and low resolution) networks architecture, which can efficiently leverage high-resolution data and directly generate high-resolution depth map. This new architecture effectively addresses the above problems and significantly improves the performance. The generated depth maps are fairly sharp and handle the slim objects and distant objects well. 

As mentioned earlier, non-texture region is also an intractable problem for unsupervised depth learning. Supervisory signal is only derived from the image area that has varying intensity, meaning that the loss in the interior of low-texture region is always very scarce which may result in black hole in the depth map. Thus we propose a Self-assembled Attention module combined with dual networks to handle the non-texture problem. This module also leads to considerable improvement.

Empirical evaluation on the KITTI~\cite{geiger2012we} and Make3D~\cite{saxena2006learning,saxena2009make3d} benchmarks demonstrates that our approach outperforms existing state-of-the-art methods for the monocular depth estimation task.

\vspace{-1mm}
\begin{figure*}[t]
\begin{center}
   \includegraphics[width=1\linewidth]{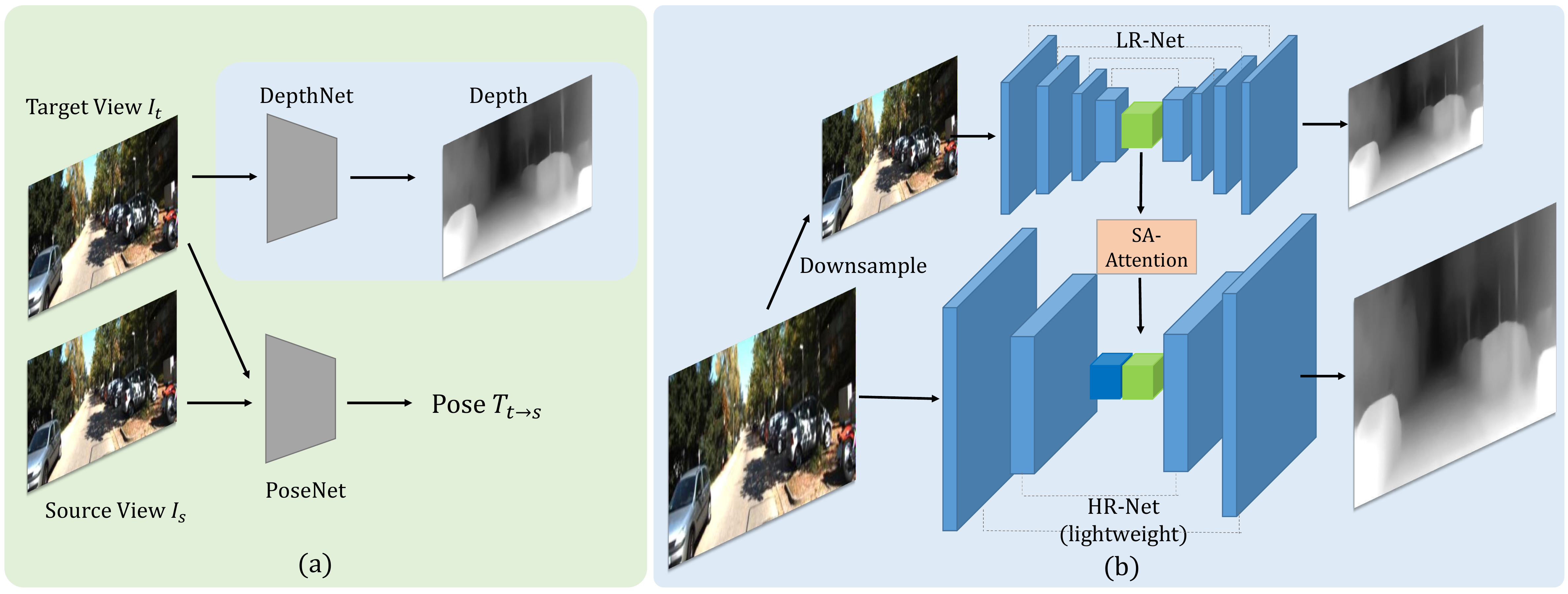}
\end{center}
   \vspace{-3mm}
   \caption{Overview of our method. (a) is the pipeline of unsupervised depth learning, which consists of depth network and pose network. (b) is the specific architecture of DepthNet in (a), which is composed of LR-Net, HR-Net and SA-Attention module. LR-Net takes low-resolution images as input and HR-Net takes high-resolution images as input. The deepest global features in LR-Net are fed to HR-Net. SA-Attention module refines the information flowing from LR-Net to HR-Net.}
\label{fig:overview}
\end{figure*}

\section{Related Work}

Here we review the works which take single view image as input and estimate the depth of each pixel. These works are classified into supervised depth learning and unsupervised depth learning. We also introduce some super resolution and self-attention works that are related to our work.

\textbf{Supervised depth learning}\quad Supervised depth learning trains the model with RGB image and ground truth depth label where there is clear supervision for each pixel. These methods need ground truth labels captured with time of flight device~\cite{heide2013low} or RGBD cameras, which has the limitation of high cost or limited depth range. ~\cite{saxena2009make3d} uses Markov Random Field (MRF) to infer a set of “plane parameters” for the given scenes.~\cite{karsch2014depth} tries non-parametric sampling to estimate the depth by matching the given image with the images in depth dataset. Eigen et al.~\cite{eigen2014depth} are the first to employ CNN in learning depth and they also proposed to use two networks. But there exists significant difference between theirs and our architecture. Their motivation is that the coarse depth map generated by a coarse-scale network can be used as additional information and concatenated with RGB images as the input of another network. Both networks take the same low-resolution images as inputs and the resolution of outputs is smaller. Our dual networks can directly process high-resolution images and output high-resolution depth maps.~\cite{laina2016deeper} builds a deep fully convolutional network based on ResNet~\cite{he2016deep}. A significant gain is achieved by leveraging the powerful deep learning method.~\cite{xu2017multi,liu2018planenet} try to explore the structured information in the depth map by either the CRF or explicit plane models.~\cite{fu2018deep} treats depth estimation as a classification problem to achieve robust inference. Other works find that depth estimation can be combined with other tasks and benefit from each other, \eg, normal~\cite{eigen2015predicting,qi2018geonet}, segmentation~\cite{liu2010single,ladicky2014pulling} and optical flow~\cite{mayer2016large}. Due to lack of large scale depth labels,  ~\cite{kundu2018adadepth} increases the performance of depth learning with synthetic data and GAN~\cite{goodfellow2014generative} loss.~\cite{li2018megadepth} uses multi-view Internet photo collections to generate training data. ~\cite{yang2018deep} uses sparse points whose depth are estimated from SLAM system as supervision. 

\textbf{Unsupervised depth learning}\quad 
There is no ground truth depth label for self-supervised depth learning. Instead the network training is supervised by multiview constraint, \eg, multiview images captured by stereo~\cite{godard2017unsupervised,li2018undeepvo} or multiview cameras or video captured by monocular cameras~\cite{zhou2017unsupervised}. An appearance loss is calculated by warping the source view to the target view using the inferred depth.~\cite{mahjourian2018unsupervised} adds a consistency constraint between the estimated depth from different views.~\cite{yang2018lego} combines depth learning with normal and edge extraction and achieves a better performance.~\cite{zou2018df} proposes to leverage optical flow to estimate occlusions and remove invalid regions in computing the loss.~\cite{atapour2018real} utilizes synthetic data and style transfer~\cite{zhu2017unpaired} to collect more diverse training data.~\cite{li2018undeepvo} uses stereo image pairs to recover the scale.

\textbf{CNN for Super Resolution}\quad Deep-learning-based Super Resolution (SR) is also a popular research topic recently, which aims to recover a high-resolution image from low-resolution image. Dong et al.~\cite{dong2014learning} was the first to apply deep learning method in SR. Huang et al.~\cite{huang2015single}  extended
self-similarity based SR and used the detected perspective geometry to guide the patch searching. Kim et al.~\cite{kim2016accurate} used VGG as backbone and achieved better performance. Shi et al.~\cite{shi2016real} proposed an efficient sub-pixel convolution layer which learns an array of upscaling filters to upscale the LR feature maps. The architecture of SR is also used in unsupervised depth learning by Pillai et al.~\cite{pillai2018superdepth}. But only using low-resolution images as input is not able to fully explore the valuable information contained in high-resolution images. Our dual networks are different from super resolution and able to process high-resolution data efficiently. Especially for unsupervised depth learning, the resolution of training data significantly influences the performance. 

\textbf{Self-Attention}\quad Self-attention mechanism~\cite{vaswani2017attention,bahdanau2014neural,yang2016stacked,li2018_zoom_journal} calculates the response at a position in a sequence by attending to all positions within the same sequence. In neural network, the convolutional and recurrent operations can only process on local neighborhood at a time.~\cite{wang2018nonlocal} presents non-local operations to capture long-range dependencies in the whole input image.~\cite{zhang2018self} learns to efficiently find global, long-range dependencies within internal representations of images.~\cite{malinowski2018learning} uses a non-local mechanism for hard attention on visual question answering datasets. Our work aims to leverage self-attention as a mechanism to enhance supervisory signal in non-texture regions and refine the global features.

\section{Approach}

In this section, we first review the nature of 3D geometry of unsupervised depth learning. Then we introduce the proposed dual networks architecture and SA-Attention module.

\subsection{Framework}
\label{sec31}
The fundamental idea behind unsupervised depth learning is multiview geometry constraint. Given two frames $I_{s}$ and $I_t$ with known camera intrinsics, once the relative pose $T_{t\xrightarrow{}s}$ and the scene depth of $I_t$ are estimated, we can synthesize $I_t$ from $I_{s}$ as:
\begin{equation}
    p_{s\xrightarrow{}t} = KT_{t\xrightarrow{}s}D_t(p_t)K^{-1}p_{t}
    \label{equ:flow}
\end{equation}
where $p_t$ and $p_{s\xrightarrow{}t}$ denote the homogeneous coordinates of a pixel in $I_{t}$ and the synthesized view $I_{s\xrightarrow{}t}$ respectively, $D_t$ denotes the depth of target view and K denotes the camera intrinsic matrix. Then the homogeneous coordinates $p_{s\xrightarrow{}t}$ can be projected to the image plane in a fully differential manner~\cite{jaderberg2015spatial} to obtain the synthesized image $I_{s\xrightarrow{}t}$.\par
The entire pipeline consists of two main modules: the DepthNet and the PoseNet, which aim to estimate monocular depth and pose between nearby views respectively. The supervisory loss is from the appearances difference between the target view and synthesized views as:
\begin{equation}
    L_{vs} = \sum_{j}\left |{I_t(j) - I_{s\xrightarrow{}t}(j) }\right |
\end{equation}
where $j$ indexes the pixel coordinates.

\subsection{Dual Network Architecture}

\textbf{Motivation}\quad As mentioned above, due to the special nature of unsupervised depth learning, the resolution of training images significantly impacts the training effect. The deficiencies of training with low-resolution images can be summarized as below:

1) Due to the decrease of resolution, the accuracy of disparity deteriorates which results in large error on distant objects.

2) Upsampling the low-resolution depth map to high-resolution blurs the border of objects.

3) Loss of fine-grained details makes the model prone to ignore slim objects like traffic pole and tree.

However, directly training with high-resolution images requires vast computation resource. On the other hand, reducing the scale of the model will also deteriorate the performance. To address this problem, we propose a dual networks architecture which can fully explore the rich information contained in high resolution data while avoiding expensive cost. The key idea is that low-resolution data already contain enough global semantic information, and at the same time the most valuable information of high-resolution data are the fine-grained details. It is difficult and unnecessary to train high-resolution data in a single network. This means that we can separate the process of feature extraction into two parts. The first part captures global semantic information and the second part fills in the details.

\textbf{Design}\quad As shown in Figure~\ref{fig:overview}, the dual networks architecture consists of three components: Low-Resolution Network (LR-Net), High-Resolution Network (HR-Net) and SA-Attention module that links these two networks. LR-Net takes low-resolution ($128\times 416$) images as input and HR-Net takes high-resolution ($384\times 1248$) images as input. Both networks share similar encoder-decoder architecture with skip-connection and generate low-resolution and high-resolution depth maps respectively. Their supervisory signal comes from the photometric loss computed by predicted depth and pose. LR-Net is designed to extract the important global semantic features from low-resolution images so it contains more convolutional layers and more parameters. However the training of LR-Net is efficient due to its small input size.

HR-Net is designed as a lightweight and shallow model so it only adds a small amount of overhead while it directly processes high-resolution data. Due to its limited capacity, HR-Net is not able to generate plausible result by itself. This module is specifically used to extract the fine-grained details in high-resolution images which will be combined with global features passed by LR-Net to generate high-resolution and high-accuracy depth map. More specifically, the deepest features in LR-Net with smallest size is delivered to HR-Net and concatenated with the deepest features in HR-Net. Then the concatenated features are upsampled gradually in the decoder and the details are also filled in. More details are shown in supplementary material. In addition, since the photometric loss can be computed by high-resolution depth map, the supervisory signal is more accurate which improves the performance.

In practice, both networks adopt a multi-scale architecture and predict depth maps with $\frac{1}{1}, \frac{1}{2}, \frac{1}{4}, \frac{1}{8}$ resolution relative to the input size. LR-Net is firstly trained to achieve a considerable performance. Then the parameters are fixed and used as global features extractor for training HR-Net. Quantitative results of these two networks are shown in the ablation study later. 

\subsection{SA-Attention Module}

Although with the help of dual networks, our model already achieves state-of-the-art performance, there still exists some problem in this pipeline. Since the important global features are learned by LR-Net, the accuracy of the learned global features directly affects the prediction of HR-Net. In some cases where severe non-texture region exists, the global features of that region cause much deviation and it is difficult for HR-Net to rectify that error.

To alleviate the non-texture problem, we propose a Self-assembled Attention module to refine the global features before they are fed to HR-Net as shown in Figure~\ref{fig:overview}. The key idea is that two pixels with similar features, appearance or close spatial distance are more likely to have similar depth. SA-Attention module explicitly calculates the similarity between each position and all the other positions, and similar pixels are bundled together. In the SA module, we concatenate the input features, RGB values of resized input image and the pixels' coordinates together in feature channel. Then the concatenated feature is embedded in a low-dimensional space. Suppose the input feature is $f_i$, this operation can be done by a conv layer to obtain embedded feature $f'_{i}$. The assembled feature at the $j$ position $f''_{j}$ can be written as a weighted sum of other similar features:

\begin{equation}
  f''_{j}=\sum_{i}w_{ij}\cdot f'_i
\end{equation}
where $i$ indexes the pixel coordinates and $w_{ij}$ represents the similarity between feature at position $i$ and $j$. There exist several choices to evaluate the similarity of two vectors, such as L1, L2 distance or cosine similarity. In practice, we use dot product of vectors for its simplicity:

\begin{equation}
  w_{ij}=f'_{i}\cdot f'_j
\end{equation}
Then the assembled feature $f''_{j}$ is passed to HR-Net. This module serves like a valve and refines the information flowing from LR-Net to HR-Net since it ensures pixels with strong similarity to have the similar depth features, which enhances the supervisory signal in non-texture regions. Its quantitative effect is shown in the ablation study later.

\begin{figure*}
\begin{center}
  \includegraphics[width=1\linewidth]{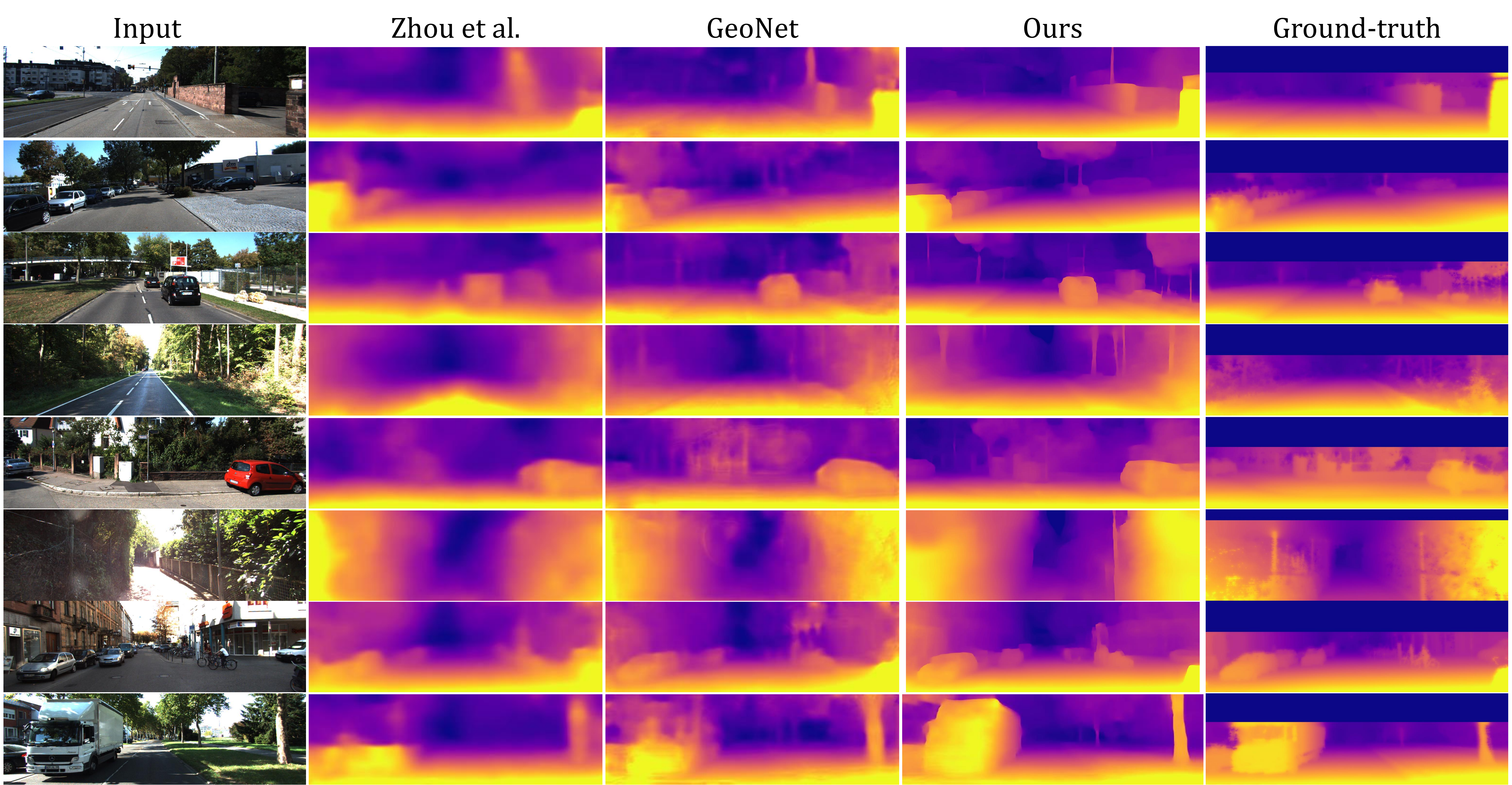}
\end{center}
   \vspace{-3mm}
   \caption{Qualitative comparison between Zhou~\etal~\cite{zhou2017unsupervised}, GeoNet~\cite{yin2018geonet}, ours and ground-truth (interpolated for visualization). In some cases, our results are more accurate than the depth maps obtained by laser scanner (\eg, the car window in the last row) since the laser can not handle transparent objects like glass window and returns the depth value behind the glass.}
\label{fig:kitti_quqa}
\end{figure*}

\subsection{Loss}
In this subsection we introduce the components of our training loss function. \par
\textbf{Photometric Loss}\quad Following~\cite{godard2017unsupervised}, we adopt a combination of L1 and the Structural Similarity (SSIM)~\cite{wang2004image} for  appropriate assessment of the discrepancy between two images. In addition, per-pixel minimum trick proposed by~\cite{godard2018digging} is also adopted. This trick is that we use three views to compute the photometric loss: one target view and two source views. So we obtain two error maps from two source views. Instead of averaging both error maps we calculate
their minimum. This is an effective way to handle occlusion/disocclusion. So the final photometric loss is
\begin{equation}
    L_{ph}= \sum_{p}\min_{s}(\alpha L_{SSIM}+(1-\alpha)\Vert{I_t(p) - I_{s\xrightarrow{}t}(p) }\Vert_1)
    \label{equ:lSSMI2}
\end{equation}
where $p$ indexes over pixel coordinates, $s$ denotes the index of source views, $\alpha$ is set to 0.85 and $L_{SSIM}$ represents
\begin{equation}
    L_{SSIM}=\frac{1-SSIM\left( I_t, I_{s\xrightarrow{}t} \right)}{2}
    \label{equ:lSSMI}
\end{equation}

\textbf{Smoothness Loss}\quad Besides photometric loss, edge-aware
depth smoothness loss is adopted which encourages the network to generate smooth prediction in continuous region while preserving sharp edge in discontinuous region:
\begin{equation}
    L_{smooth}=\sum_{p}{\left|{\nabla}D_t(p)\right|\cdot \left(e^{-\left|{\nabla}I_t(p)\right|}\right)^T}
    \label{equ:lsmooth}
\end{equation}
where $\nabla$ is the vector differential operator, and $T$ denotes the transpose of image gradient weighting.

\textbf{Feature Reconstruction Loss}\quad Even though only photometric loss and smoothness loss are already sufficient to obtain a comparable result, this supervision is still not accurate. 
In practice, we adopt the feature reconstruction loss proposed by~\cite{zhan2018unsupervised}. The homogeneous coordinates computed by Equation~\ref{equ:flow} can also be used to warp the last decoder layer's features which contain 16 channels in source view to the target view. Then L2 norm is computed with respect to the target view's features $f_t$ and synthesized view's features $f_{s\xrightarrow{}t}$. Again the per-pixel mimimum trick is also used here:
\begin{equation}
    L_{fr}=\sum_{p}\min_{s}\Vert{f_t(p) - f_{s\xrightarrow{}t}(p) }\Vert_2
    \label{equ:fr}
\end{equation}

Then the total loss for training LR-Net is the combination of the above three losses:
\begin{equation}
    L_{total}=\lambda_{1}L_{ph}+\lambda_{2}L_{smooth}+\lambda_{3}L_{fr}
    \label{equ:total}
\end{equation}

As for the training of HR-Net, we only use $L_{ph}$ and $L_{smooth}$ since the cost of computing Feature Reconstruction Loss on high-resolution feature maps is expensive.

\section{Experiments}

In this section, we evaluate our approach on KITTI dataset~\cite{geiger2012we} and Make3D dataset~\cite{saxena2006learning,saxena2009make3d}. Camera pose evaluation, ablation study of each module and visualization of the results are also conducted. Finally the implementation details are clarified.

\begin{table*}[htbp]
\vspace{2mm}
\renewcommand\tabcolsep{4.0pt}
  \centering
  \resizebox{1.0 \linewidth}{!}{
    \begin{tabular}{ccccccccccc}
    \toprule
    \multicolumn{2}{c}{\multirow{2}[4]{*}{Method}} & \multirow{2}[4]{*}{Dataset} & \multirow{2}[4]{*}{Supervision} & \multicolumn{4}{c}{Error metric} & \multicolumn{3}{c}{Accuracy metric} \\
\cmidrule(l){5-8} \cmidrule(l){9-11}   \multicolumn{2}{c}{} &       &       & \multicolumn{1}{l}{Abs Rel} & \multicolumn{1}{l}{Sq Rel} & \multicolumn{1}{l}{RMSE} & \multicolumn{1}{l}{RMSE log} & \multicolumn{1}{l}{$\delta<1.25$} & \multicolumn{1}{l}{$\delta<1.25^2$} & \multicolumn{1}{l}{$\delta<1.25^3$} \\
    \midrule
    \multicolumn{2}{c}{Train set mean} & -     & -     & 0.361 & 4.826 & 8.102 & 0.377 & 0.638 & 0.804 & 0.894 \\
    \multicolumn{2}{c}{Eigen et al.~\cite{eigen2014depth}} & K     & Depth & 0.203 & 1.548 & 6.307 & 0.282 & 0.702 & 0.890 & 0.890 \\
    \multicolumn{2}{c}{Liu et al.~\cite{liu2014discrete}} & K     & Depth & 0.201 & 1.584 & 6.471 & 0.273 & 0.680 & 0.898 & 0.967 \\
    \multicolumn{2}{c}{Godard et al.~\cite{godard2017unsupervised}} & K     & Stereo & 0.148 & 1.344 & 5.927 & 0.247 & 0.803 & 0.922 & 0.964 \\
    \multicolumn{2}{c}{Godard et al.~\cite{godard2017unsupervised}} & K+CS  & Stereo & 0.124 & 1.076 & 5.311 & 0.219 & 0.847 & 0.942 & 0.973 \\
    \multicolumn{2}{c}{Kuznietsov et al.~\cite{kuznietsov2017semi}} & K     & Depth+Stereo & 0.113 & 0.741 & 4.621 & 0.189 & 0.862 & 0.960 & 0.986 \\
    \multicolumn{2}{c}{DORN~\cite{fu2018deep}} & K     & Depth & 0.072 & 0.307 & 2.727 & 0.120 & 0.932 & 0.984 & 0.994 \\
    \multicolumn{2}{c}{Yang et al.~\cite{yang2018every}} & K+CS   & Stereo & 0.114 & 1.074 & 5.836 & 0.208 & 0.856 & 0.939 & 0.976 \\
    \multicolumn{2}{c}{Casser et al.~\cite{casser2019struct2depth}} & K     & Instance Label & 0.109 & 0.825 & 4.750 & 0.187 & 0.874 & 0.958 & 0.983 \\
    \midrule
    \multicolumn{2}{c}{Zhou et al.~\cite{zhou2017unsupervised}} & K     &    -   & 0.183 & 1.595 & 6.709 & 0.270 & 0.734 & 0.902 & 0.959 \\
    \multicolumn{2}{c}{GeoNet~\cite{yin2018geonet}} & K     &   -    & 0.155 & 1.296 & 5.857 & 0.233 & 0.793 & 0.931 & 0.973 \\
    \multicolumn{2}{c}{DDVO~\cite{wang2018learning}} & K     &    -   & 0.151 & 1.257 & 5.583 & 0.228 & 0.810 & 0.936 & 0.974 \\
    \multicolumn{2}{c}{Godard et al.(V2)~\cite{godard2018digging}} & K     &    -   & 0.129 & 1.112 & 5.180  & 0.205 & 0.851 & 0.952 & 0.978 \\
    \multicolumn{2}{c}{Yang et al.~\cite{yang2018every}} & K     &    -   & 0.131 & 1.254 & 6.117  & 0.220 & 0.826 & 0.931 & 0.973 \\
    \midrule
    \multicolumn{2}{c}{\textbf{Ours}} & K     &   -    & \textbf{0.121} & \textbf{0.837} & \textbf{4.945} & \textbf{0.197} & \textbf{0.853} & \textbf{0.955} & \textbf{0.982} \\
    \bottomrule
    \end{tabular}%
    }
    \vspace{2mm}
    \caption{Comparison to existing methods on KITTI 2015~\cite{geiger2012we} using the Eigen split~\cite{eigen2014depth}. Our method achieves state-of-the-art result.}
  \label{tab:KITTI}%
\end{table*}%

\subsection{KITTI Depth}

We evaluate our approach on KITTI 2015~\cite{Menze2015ISA} by the split of Eigen et al.~\cite{eigen2014depth}. As shown in Table~\ref{tab:KITTI}, our results significantly outperform existing state-of-the-art methods. Qualitative comparisons are shown in Fig~\ref{fig:kitti_quqa}. Visually our depth predictions are sharper and unambiguous. Slim structures like traffic poles and trees are also handled well. In some cases, such as the glass windows on the cars, our model performs better than the measurement obtained by laser radar. In addition, the output sizes of previous methods are mostly $128\times 416$ or $192\times 640$, but our model's output size is $384\times 1248$, which is close to KITTI's original size. The static frames are removed as in~\cite{zhou2017unsupervised} during training since the supervisory signal comes from the disparity generated by the camera motion. Median scaling proposed by~\cite{eigen2014depth} is adopted to align the predictions with the ground-truth depth. All the settings during evaluation are the same as previous methods~\cite{zhou2017unsupervised,yin2018geonet}. \\

\subsection{Make3D}

\begin{table}[h]
\renewcommand\tabcolsep{3pt}
  \centering
  \resizebox{1.0 \linewidth}{!}{
    \begin{tabular}{ccccccc}
    \toprule
    \multicolumn{2}{c}{\multirow{2}[4]{*}{Method}} & \multicolumn{2}{c}{Supervision} & \multirow{2}[4]{*}{Abs Rel} & \multirow{2}[4]{*}{Sq Rel} & \multirow{2}[4]{*}{RMSE} \\
\cmidrule{3-4}    \multicolumn{2}{c}{} & \multicolumn{1}{l}{Depth} & \multicolumn{1}{c}{Pose} &  &       &  \\
    \midrule
    \multicolumn{2}{c}{Train set mean} &       &       & 0.893 & 13.98 & 12.27 \\
    \multicolumn{2}{c}{Karsch et al.~\cite{karsch2014depth}} & \checkmark   &       & 0.428 & 5.079 & 8.389 \\
    \multicolumn{2}{c}{Liu et al.~\cite{liu2014discrete}} & \checkmark   &       & 0.475 & 6.562 & 10.05 \\
    \multicolumn{2}{c}{Laina et al.~\cite{laina2016deeper}} & \checkmark   &       & 0.204 & 1.840 & 5.683 \\
    \multicolumn{2}{c}{Godard et al.~\cite{godard2017unsupervised}} &       & \checkmark   & 0.544 & 10.94 & 11.76 \\
    \midrule
    \multicolumn{2}{c}{Zhou et al.~\cite{zhou2017unsupervised}} &       &       & 0.383 & 5.321 & 10.47 \\
    \multicolumn{2}{c}{DDVO~\cite{wang2018learning}} &       &       & 0.387 & 4.720 & 8.090 \\
    \multicolumn{2}{c}{Godard et al.(V2)~\cite{godard2018digging}} &       &       & 0.361 & 4.170 & 7.821 \\
    \multicolumn{2}{c}{\textbf{Ours}} &       &       & \textbf{0.318} & \textbf{2.288} & \textbf{6.669} \\
    \bottomrule
    \end{tabular}%
    }
  \vspace{3mm}
  \caption{\label{tab:make3d}Evaluation on Make3D~\cite{saxena2009make3d} dataset. Our result is obtained by the model trained on KITTI without any fine-tuning.}
\end{table}%

We directly evaluate our model trained by KITTI on Make3D dataset {\it without} any fine-tuning. We just resize the test images to $384\times 1248$ resolution and feed them to the network. As shown in Table~\ref{tab:make3d}, our result outperforms existing state-of-the-art methods as well although it has not been trained on that dataset, which shows the generalization ability of our model.

\begin{table*}[htbp]
  \centering
    \begin{tabular}{cccccccc}
    \toprule
    \multirow{2}[3]{*}{Method} & \multicolumn{4}{c}{Error metric} & \multicolumn{3}{c}{Accuracy metric} \\
\cmidrule(l){2-5} \cmidrule(l){6-8}  & \multicolumn{1}{l}{Abs Rel} & \multicolumn{1}{l}{Sq Rel} & \multicolumn{1}{l}{RMSE} & \multicolumn{1}{l}{RMSE log} & \multicolumn{1}{l}{$\delta<1.25$} & \multicolumn{1}{l}{$\delta<1.25^2$} & \multicolumn{1}{l}{$\delta<1.25^3$} \\
    \midrule
    Baseline (LR-Net only) & 0.132 & 0.929 & 5.208 & 0.209 & 0.833 & 0.946 & 0.979 \\
    LR-Net+HR-Net (w/o SA module) & 0.123 & 0.881 & 5.016 & 0.198 & 0.851 & 0.955 & 0.982 \\
    LR-Net+HR-Net (w/ SA module) & 0.121 & 0.837 & 4.945 & 0.197 & 0.853 & 0.955 & 0.982 \\
    \bottomrule
    \end{tabular}%
  \vspace{2mm}
  \caption{Evaluation of each component in our model on KITTI's eigen test split.}
  \label{tab:ablation_dual}
\end{table*}%

\begin{table*}[htbp]
  \centering
    \begin{tabular}{ccccccccc}
    \toprule
    \multirow{2}[3]{*}{Method} & \multirow{2}[3]{*}{Distance} & \multicolumn{4}{c}{Error metric} & \multicolumn{3}{c}{Accuracy metric} \\
\cmidrule(l){3-6} \cmidrule(l){7-9}   &       & \multicolumn{1}{l}{Abs Rel} & \multicolumn{1}{l}{Sq Rel} & \multicolumn{1}{l}{RMSE} & \multicolumn{1}{l}{RMSE log} & \multicolumn{1}{l}{$\delta<1.25$} & \multicolumn{1}{l}{$\delta<1.25^2$} & \multicolumn{1}{l}{$\delta<1.25^3$} \\
    \midrule
    \multirow{2}[2]{*}{w/o HR-Net} & $\le 20m$     & 0.108 & 0.448 & 2.107 & 0.153 & 0.901 & 0.973 & 0.990 \\
    & $>20m$ & 0.199 & 2.685 & 9.885 & 0.310 & 0.648 & 0.865 & 0.940 \\
    \midrule
    \multirow{2}[2]{*}{w/ HR-Net} & $\le 20m$ & 0.102 & 0.391 & 1.959 & 0.146 & 0.912 & 0.977 & 0.990 \\
    & $> 20m$ & 0.187 & 2.430 & 9.695 & 0.305 & 0.662 & 0.875 & 0.946 \\
    \bottomrule
    \end{tabular}%
  \vspace{2mm}
  \caption{Evaluation of the dual networks architecture on distant objects and nearby objects. The gain of distant objects (Abs Rel: 0.012) is higher than that of nearby objects (Abs Rel: 0.006).}
  \label{tab:ablation_disntance}%
\end{table*}%

\begin{table}[htbp]
  \centering
    \begin{tabular}{lllcc}
    \toprule
    \multicolumn{3}{c}{} & w/o HR-Net & w/ HR-Net \\
    \midrule
    \multicolumn{3}{c}{Score} & 20784.4 & 106987.2  \\
    \bottomrule
    \end{tabular}%
    \vspace{2mm}
  \caption{\label{tab:sharpness} Evaluation of the depth maps' average sharpness on KITTI's eigen test split. The Tenengrad score coarsely represents the sharpness of an image. Higher score means sharper boundary.}
\end{table}%

\begin{table}[htbp]
  \centering
    \begin{tabular}{lllcc}
    \toprule
    \multicolumn{3}{c}{Region} & w/o SA & w/ SA \\
    \midrule
    \multicolumn{3}{c}{Road} & 0.085 & 0.080  \\
    \multicolumn{3}{c}{Tree \& grass} & 0.190 & 0.192 \\
    \bottomrule
    \end{tabular}%
    \vspace{1mm}
  \caption{\label{tab:sa_region} Evaluation of SA module in regions with and without noticeable textures (i.e. tree \& grass vs. road) respectively  with Abs rel metric. Without SA means the global features are directly fed to the HR-Net. The gain mostly comes from low-texture regions (Abs rel: 0.005) compared with regions (Absrel: -0.002) with noticeable textures.}
  \vspace{-4mm}
\end{table}%

\begin{figure}[htbp]
\begin{center}
  \includegraphics[width=1\linewidth]{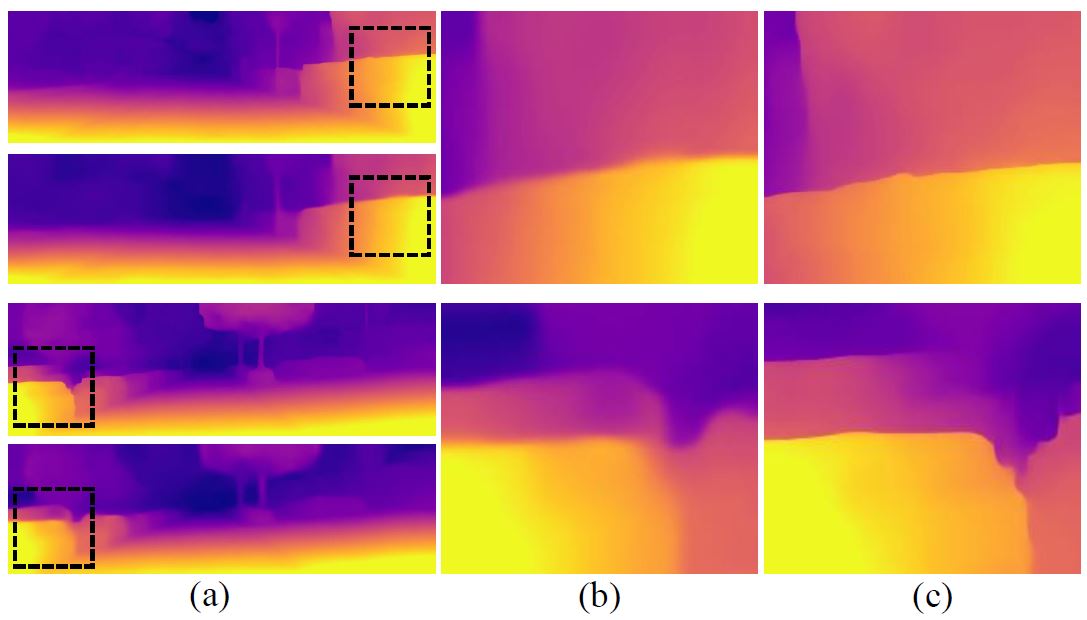}
\end{center}
   \vspace{-3mm}
   \caption{Comparison of the depth map at the object boundary. (a) Predicted depth maps with (top) and without HR-Net. (b) Zoomed patches without HR-Net. (c) Zoomed patches with HR-Net. Since HR-Net directly generates high-resolution depth map, the boundary is fairly sharp.}
\label{fig:border}
\end{figure}

\begin{figure*}[ht]
\begin{center}
  \includegraphics[width=1\linewidth]{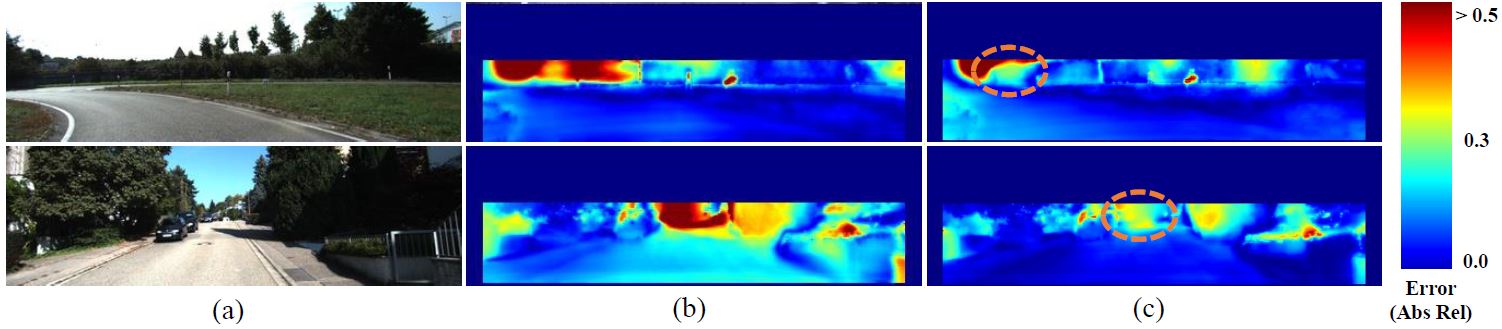}
\end{center}
   \vspace{-3mm}
   \caption{Evaluation of our dual networks' effect on distant objects. (a) Input images. (b) Error maps without HR-Net. (c) Error maps with HR-Net. The error induced by distant objects (orange circles) is improved by the dual networks visibly.}
\label{fig:distant}
\end{figure*}

\begin{figure}
\begin{center}
  \includegraphics[width=1\linewidth]{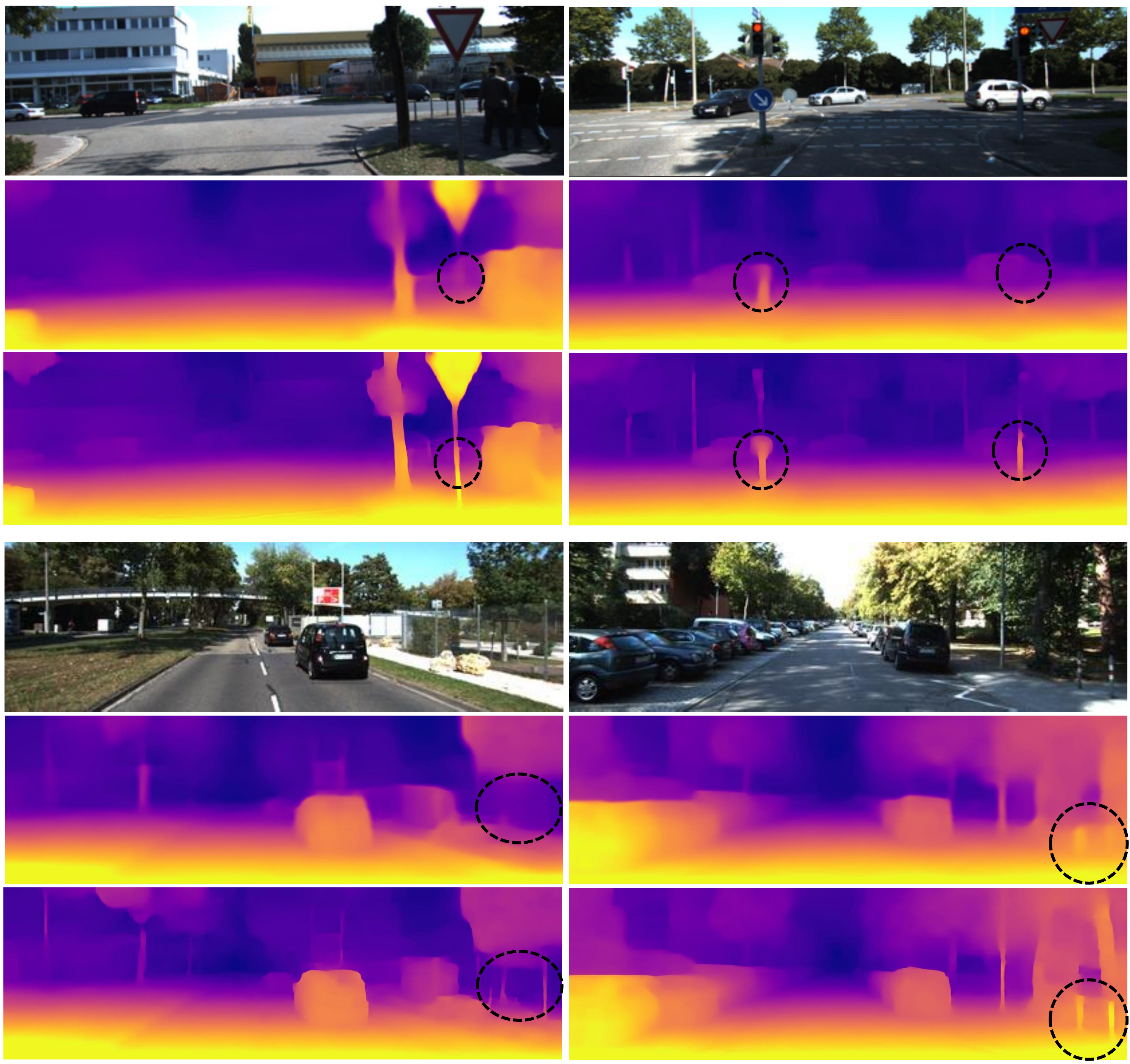}
\end{center}
   \vspace{-2mm}
   \caption{Comparison of slim objects' prediction. Top: original images. Middle: prediction without HR-Net. Bottom: prediction with HR-Net. The prediction of the traffic poles is improved significantly.}
\label{fig:pole}
\vspace{-6mm}
\end{figure}

\begin{table*}[htbp]
\centering
    \begin{tabular}{ccccccccc}
    \toprule
    \multicolumn{1}{c}{BackBones} & 
    \multicolumn{1}{c}{Params(M)} & 
    \multicolumn{1}{c}{GFLOPs} & 
    \multicolumn{1}{c}{Abs Rel} & 
    \multicolumn{1}{c}{Sq Rel} & 
    \multicolumn{1}{c}{RMSE} & 
    \multicolumn{1}{c}{$\delta<1.25$} & 
    \multicolumn{1}{c}{$\delta<1.25^2$} & 
    \multicolumn{1}{c}{$\delta<1.25^3$} \\
    \midrule
    MobileNet V2~\cite{sandler2018mobilenetv2}& 5.37 & 17.02  & 0.147   & 0.998 & 5.185 & 0.810 & 0.940 & 0.976 \\
    ResNet18 & 16.43  & 33.11  & 0.138  & 0.968 & 5.281 & 0.823 & 0.947 & 0.980 \\
    Dual-network(ResNet50) & 34.16 & 25.80 & 0.121 & 0.837 & 4.945 & 0.853 & 0.955 & 0.982\\
    \midrule
    w/ pt & - & - & 0.121 & 0.837 & 4.945 & 0.853 & 0.955 & 0.982\\
    w/o pt & - & - & 0.135 & 0.973 & 5.235 & 0.823 & 0.947 & 0.980\\
    \bottomrule
    \end{tabular}%

     \vspace{1mm}
    \caption{Comparison of our dual-network with other lightweight backbones when training with images of $384\times 1248$ resolution. Params and GFLOPs denote the number of parameters and GFLOPs of the DepthNet (including encoder and decoder). w/ pt and w/o pt are the results of dual-network with and without ImageNet pretraining.}
    \vspace{-5mm}
    \label{tab:res1}%
\end{table*}%

\subsection{KITTI Odometry}

We have evaluated the performance of our method on KITTI odometry split. As same as previous methods, we use the 00-08 sequences for training and the 09-10 sequences for testing. Our PoseNet is the same as Zhou et al.~\cite{zhou2017unsupervised}'s. As shown in Table.~\ref{tab:pose1}, the pose result of dual-network is better than LR-Net's while their PoseNets share the same architecture. The gain of the pose comes from more accurate depth prediction. ``Training frames'' means the number of input frames used by PoseNet during training and testing.

\subsection{Ablation Study}

\textbf{Dual Networks}\quad As shown in Table~\ref{tab:ablation_dual}, Figure~\ref{fig:border} and Figure~\ref{fig:pole}, both quantitative result and visual effect are improved noticeably with the help of HR-Net. The depth of traffic poles and trees are more accurate. Since the depth map is of high-resolution and need not to be upsampled, the border of objects is sharper than low-resolution result. 

\textbf{Distant Objects}\quad As mentioned before, the disparity of distant objects becomes more accurate since our HR-Net can fully explore the information of high-resolution images. Here we respectively evaluate our result on distant objects and nearby objects. We consider the pixels with more than 20m ground-truth depth as distant objects and others as nearby objects. The quantitative result is shown in Table~\ref{tab:ablation_disntance} and error map samples are shown in Figure~\ref{fig:distant}. Both of them clearly show that our dual networks effectively reduce the error produced by distant objects.

\textbf{Objects' boundary}\quad Qualitative comparison of objects' boundary is shown in Figure~\ref{fig:border}. To quantitatively evaluate the sharpness
of the depth maps generated by HR-Net, we use Tenengrad measure~\cite{krotkov2012active} to coarsely compare the sharpness of the results with and without HR-Net (shown in Table~\ref{tab:sharpness}). The definition of Tenengrad function is written as:
\vspace{-2mm}
\begin{equation}
    T=\sum_{x}\sum_{y}|G(x,y)|,( G(x,y)>t )
    \label{equ:tenengrad}
\end{equation}
\begin{equation}
    G(x,y)=\sqrt{G_x^2(x,y)+G_y^2(x,y)}
\end{equation}
where $T$ denotes Tenengrad gradient, $G_x,G_y$ denote the horizontal and vertical gradients of depth map obtained by Sobel filters, $t$ denotes the threshold and $x,y$ denote pixel coordinates.

\textbf{SA-Attention Module}\quad To verify SA module for low-texture region, we use segmentation model PSPNet~\cite{zhao2017pyramid} pretrained on Cityscapes to extract the semantic labels of KITTI's test split.  Then we evaluate the performance of SA module in regions with and without noticeable textures (i.e. tree \& grass vs. road) respectively as shown in Table~\ref{tab:sa_region}. The SA module has much higher gain in low-texture regions compared with regions with noticeable textures.

\textbf{Overhead Comparison}\quad Table~\ref{tab:res1} shows the results of training with different DepthNet encoders. Suppose the input size is $384\times 1248$, our dual-network has more parameters and lower GFLOPs than ResNet18, which means more efficient. And it also performs better than other two backbones. The lower part also reveals that ImageNet pretraining is important.

\subsection{Implementation Details}
\label{sec:impdetail}
Our model is implemented with the Tensorflow~\cite{abadi2016tensorflow} framework. Network architectures are shown in supplementary material. We first train the LR-Net for 40 epochs. Then we fix the parameters of LR-Net and continue to train the HR-Net for another 40 epochs. We use mini-batch size of 4 at both stages and optimize the network with Adam~\cite{kingma2014adam}, where $\beta_1=0.9,\beta_2=0.999$. The settings of learning rate at both stages are the same as below:
\vspace{-2mm}
\begin{equation}
lr=\left\{
\begin{array}{rcl}
0.0002 & & {epochs \leq 10}\\
0.0001 & & {10<epochs\leq 20}\\
0.00005 & & {20<epochs}\\
\end{array} \right.
\end{equation}
 Zero padding before convolution is replaced by reflection padding as in~\cite{godard2018digging}. All the predicted depth are normalized as in~\cite{wang2018learning} in order to avoid the shrinking prediction during training. We observed that previous implementations generally down-sample the images beforehand for the convenience of subsequent training. Then data augmentations (e.g., random scaling and random cropping) are applied. However, down-sampling first then random cropping later will further lose the fine-grained details. We just simply rearrange the order, i.e., applying data augmentations first and down-sampling later. This trick improves the baseline's performance.

\begin{table}[H]
\vspace{-1mm}
  \centering
   \resizebox{1.0 \linewidth}{!}{
    \begin{tabular}{c|c|c|c|c|c|c|c}
    \toprule
    \multicolumn{2}{c|}{Method} & \multicolumn{2}{c|}{Sequence 09} & \multicolumn{2}{c|}{Sequence 10} & \multicolumn{2}{c}{$\#$}\\
    \midrule
    \multicolumn{2}{c|}{ORB-Slam(full)} & \multicolumn{2}{c|}{0.014 $\pm$ 0.008} & \multicolumn{2}{c|}{0.012 $\pm$ 0.011}& \multicolumn{2}{c}{-} \\
    \midrule
    \multicolumn{2}{c|}{ORB-Slam(short)} & \multicolumn{2}{c|}{0.064 $\pm$ 0.141} & \multicolumn{2}{c|}{0.064 $\pm$ 0.130} & \multicolumn{2}{c}{-}\\
    \multicolumn{2}{c|}{Zhou et al.~\cite{zhou2017unsupervised}} & \multicolumn{2}{c|}{0.021 $\pm$ 0.017} & \multicolumn{2}{c|}{0.020$\pm$ 0.015} & \multicolumn{2}{c}{5}\\
    \multicolumn{2}{c|}{GeoNet~\cite{yin2018geonet}} & \multicolumn{2}{c|}{\textbf{0.012}$\pm$ \textbf{0.007}} & \multicolumn{2}{c|}{\textbf{0.012} $\pm$ \textbf{0.009}} & \multicolumn{2}{c}{5}\\
    \multicolumn{2}{c|}{DF-Net~\cite{zou2018df}} & \multicolumn{2}{c|}{0.017$\pm$ 0.007} & \multicolumn{2}{c|}{0.015$\pm$ 0.009}  & \multicolumn{2}{c}{5} \\
    \multicolumn{2}{c|}{DDVO~\cite{wang2018learning}} & \multicolumn{2}{c|}{0.045$\pm$ 0.108} & \multicolumn{2}{c|}{0.033$\pm$ 0.074}  & \multicolumn{2}{c}{3} \\
    \midrule
    \multicolumn{2}{c|}{Ours(LR-Net)} & \multicolumn{2}{c|}{0.017$\pm$ 0.008} & \multicolumn{2}{c|}{0.016 $\pm$ 0.009}& \multicolumn{2}{c}{3}\\
    \multicolumn{2}{c|}{Ours(LR-Net+HR-Net)} & \multicolumn{2}{c|}{0.015$\pm$ 0.007} & \multicolumn{2}{c|}{0.015$\pm$ 0.009}& \multicolumn{2}{c}{3}\\
    \bottomrule
    \end{tabular}
	}
  \vspace{-1mm}
  \caption{Evaluation results of the average absolute trajectory error and standard deviation in meters. $\#$ denotes the number of training frames.}
  \label{tab:pose1}
  \vspace{-4mm}
\end{table}

\section{Conclusion}

In this paper, we propose a dual networks architecture, which consists of LR-Net and HR-Net. This architecture is able to fully explore the fine-grained details contained in high-resolution images and directly generates high-resolution depth map. The proposed techniques in this paper can
also be applied in other resolution-sensitive tasks like unsupervised flow learning, etc. However, there still exist some intractable problems in our approach like dynamic cars, pedestrians, bicycles and reflective objects, which are left to be addressed in the future work.

{\small
\bibliographystyle{ieee_fullname}
\bibliography{egbib}
}

\end{document}